\RequirePackage{snapshot}

\documentclass{article}

\usepackage{microtype}
\usepackage{graphicx}
\usepackage{subfigure}
\usepackage{booktabs}
\usepackage{amsmath}
\usepackage{amssymb}
\usepackage{cancel}
\usepackage{custom}

\usepackage{hyperref}

\usepackage{cleveref}

\usepackage[accepted]{icml2020}

\icmltitlerunning{CEB Improves Model Robustness}

\title{CEB Improves Model Robustness}

\author{%
Ian Fischer \\
Google Research \\
\texttt{iansf@google.com} \\
\and
Alexander A. Alemi \\
Google Research \\
\texttt{alemi@google.com} \\
}

\begin{document}

\twocolumn[
\maketitle
\icmlkeywords{Information Bottleneck, Robustness, Adversarial Robustness, CEB, Machine Learning}
\vskip 0.0in
]
\begin{abstract}
We demonstrate that the Conditional Entropy Bottleneck (CEB) can improve
model robustness. CEB is an easy strategy to implement and
works in tandem with  data augmentation procedures.
We report results of a large scale adversarial robustness study on
CIFAR-10, as well as the \imagenetc\ Common Corruptions Benchmark, \imageneta, and PGD attacks.
\end{abstract}

\section{Introduction}

We aim to learn models that make meaningful predictions beyond the data they
were trained on.
Generally we want our models to be \emph{robust}.
Broadly, robustness is the ability of a model to continue
making valid predictions
as the distribution the model
is tested on moves away
from the empirical training set distribution.
The most commonly reported robustness metric is
simply test set performance,
where we verify that our model continues to make valid predictions
on what we hope represents valid draws from the same data generating procedure as the training set.

Adversarial attacks
test robustness in a worst case setting, where an attacker~\citep{szegedy2013intriguing} makes limited targeted
modifications to the input that are as fooling as possible.
Many adversarial attacks have been proposed and studied (e.g.,~\citet{adversaries,carliniwagner,carlini2017adversarial,kurakin2016adversarial,madry2017towards}).
Most machine-learned systems appear to be vulnerable to adversarial examples.
Many defenses have been proposed, but few have demonstrated robustness against a powerful, general-purpose adversary~\citep{carlini2017adversarial,athalye2018obfuscated}.
Recent discussions have emphasized the need to consider forms of robustness besides adversarial~\citep{distill}.
The Common Corruptions Benchmark~\citep{hendrycks2019benchmarking} measures image models' robustness to more mild real-world perturbations.
Even these modest perturbations can fool traditional architectures.

One general-purpose strategy that has been shown to improve 
model robustness is data augmentation~\citep{autoaugment,patchgauss,fourier}.
Intuitively, by performing modifications of the inputs at training time, the model is
prevented from being too sensitive to particular features of the inputs that
don't survive the augmentation procedure.
We would like to identify complementary techniques for further improving robustness.

One approach is to try to make our models more robust by making them less sensitive to the inputs in the first place.
The goal of this work is to experimentally investigate whether, by systematically limiting
the complexity of the extracted representation using the Conditional Entropy Bottleneck (CEB), we can make our models more robust in all three
of these senses: test set generalization (e.g., classification accuracy on ``clean'' test inputs), worst-case robustness, and typical-case robustness.

\subsection{Contributions}

This paper is primarily empirical.
We demonstrate:
\begin{itemize}[leftmargin=1em]
\item CEB models are easy to implement and train.
\item CEB models show improved generalization performance over deterministic
baselines on \cifar\ and \imagenet.
\item CEB models show improved robustness to untargeted Projected Gradient Descent (PGD) attacks on \cifar.
\item CEB models trained on \imagenet\ show improved robustness on the \imagenetc\ Common Corruptions Benchmark, the \imageneta\ Benchmark, and targeted PGD attacks.
\end{itemize}

We also show that adversarially-trained models \textit{fail} to generalize to attacks they weren't trained on, by comparing the results on L$_2$ PGD attacks from~\citet{madry2017towards} to our results on the same baseline architecture.
This result underscores the importance of finding ways to make models robust that do not rely on knowing the form of the attack ahead of time.
Finally, for readers who are curious about theoretical and philosophical perspectives that may give insights into why CEB improves robustness, we recommend~\citet{ceb}, which introduced CEB, as well as~\citet{achille} and \citet{achille2018information}.

\section{Background}

\subsection{Information Bottlenecks}
The Information Bottleneck (IB) objective~\citep{tishby2000information} aims to learn a
stochastic representation $Z \sim p(z|x)$ that
retains as much information about
a target variable $Y$ while being as
compressed as possible.
The objective:\footnote{
    The IB objective is ordinarily written with
    a Lagrange multiplier $\beta \equiv \sigma(-\rho)$
    with a natural range from 0 to 1.
    Here we use the sigmoid function: $\sigma(-\rho) \equiv \frac 1 {1 + e^{\rho}}$
    to reparameterize in terms of a control parameter $\rho$ on the whole
    real line.
    As $\rho\to\infty$ the bottleneck turns off.
}
\begin{equation}
    IB \equiv \max_Z I(Z;Y) - \sigma(-\rho) I(Z;X),
    \label{eqn:ib}
\end{equation}
uses a Lagrange multiplier $\sigma(-\rho)$ to trade off
between the relevant information $(I(Z;Y))$ and
 complexity of the
 representation $(I(Z;X))$.
Because $Z$ depends only on $X$ ($Z \leftarrow X \leftrightarrow Y$), $Z$ and $Y$ are independent given $X$:
\begin{align}
I(Z;X,Y) =& I(Z;X) + \cancel{I(Z;Y|X)} \nonumber \\
=& I(Z;Y) + I(Z;X|Y).
\end{align}

This allows us to write the information bottleneck of \Cref{eqn:ib}
in an equivalent form:
\begin{equation}
    \max_Z I(Z;Y) - e^{-\rho} I(Z;X|Y).
    \label{eqn:rewrite}
\end{equation}
Just as the original Information Bottleneck objective (\Cref{eqn:ib})
admits a natural variational lower bound~\citep{vib}, so does this form.
We can variationally lower bound the mutual information
between our representation and the targets
with a variational decoder $q(y|z)$:
\begin{align}
I(Z;Y) =& \mathbb{E}_{p(x,y)p(z|x)} \left[ \log \frac{p(y|z)}{p(y)} \right] \nonumber \\
\geq& H(Y) + \mathbb{E}_{p(x,y) p(z|x)} \left[ \log q(y|z) \right].
\end{align}
While we may not know $H(Y)$ exactly for real world
datasets, in the information bottleneck formulation it
is a constant outside of our control and so can be dropped in our objective.
We can variationally upper bound our residual
information:
\begin{align}
    I(Z;X|Y) =& \mathbb{E}_{p(x,y) p(z|x)} \left[ \log \frac{p(z|x,y)}{p(z|y)} \right] \nonumber \\
        \leq& \mathbb{E}_{p(x,y)p(z|x)}\left[ \log \frac{p(z|x)}{q(z|y)} \right],
    \label{eqn:upper}
\end{align}
with a variational class conditional marginal $q(z|y)$
that approximates $\int dx\, p(z|x) p(x|y)$.
Putting both bounds together gives us
the Conditional Entropy Bottleneck objective~\citep{ceb}:
\begin{equation}
    \min_{p(z|x)} \mathbb{E}_{p(x,y)p(z|x)}
    \left[ -\log q(y|z) + e^{-\rho} \log \frac{p(z|x)}{q(z|y)}\right]
    \label{eqn:ceb}
\end{equation}
Compare this with the Variational Information Bottleneck (VIB) objective~\citep{vib}:
\begin{equation}
    \min_{p(z|x)} \mathbb{E}_{p(x,y)p(z|x)}
    \left[ \log q(y|z) - \sigma(-\rho) \log \frac{p(z|x)}{q(z)}\right].
    \label{eqn:vib}
\end{equation}
The difference between CEB and VIB is the presence
of a class conditional versus unconditional variational
marginal.
As can be seen in \Cref{eqn:upper},
using an unconditional marginal provides
a looser variational upper bound on $I(Z;X|Y)$.
CEB~(\Cref{eqn:ceb}) can be thought of
as a tighter variational approximation than
VIB~(\Cref{eqn:vib}) to \Cref{eqn:rewrite}.
Since \Cref{eqn:rewrite} is equivalent to the IB objective (\Cref{eqn:ib}),
CEB can be thought of as a tighter variational approximation to the IB
objective than VIB.

\subsection{Implementing a CEB Model}
\label{sec:implementing}
In practice, turning an existing classifier architecture into a CEB model is very simple.
For the stochastic representation $p(z|x)$ we simply use the original architecture, replacing
the final softmax layer with a dense layer with $d$ outputs.  These outputs are then
used to specify the means of a $d$-dimensional Gaussian distribution with unit diagonal
covariance.  That is, to form the stochastic representation,
independent standard normal
noise is simply added to the output of the network ($z = x + \epsilon$).
For every input, this stochastic encoder will generate a random
$d$-dimensional output vector.
For the variational classifier $q(y|z)$ any classifier network
can be used, including just a linear softmax classifier as done in these experiments.
For the variational conditional marginal $q(z|y)$ it helps to use the same distribution
as output by the classifier.
For the simple unit variance Gaussian encoding we used in these experiments,
this requires learning just $d$ parameters per class.
For ease of implementation, this can be represented as single dense linear layer mapping
from a one-hot representation of the labels to the $d$-dimensional output, interpreted as
the mean of the corresponding class marginal.

In this setup the CEB loss takes a particularly simple form:
\begin{align}
&\mathbb{E}\Bigg[ w_{y} \cdot (f(x) + \epsilon) - \log \sum_{y'} e^{w_{y'} \cdot (f(x) + \epsilon)} \label{eqn:softmax} \\
&\quad~~ - \frac{e^{-\rho}}{2}  (f(x) - \mu_{y}) \left( f(x)  - \mu_{y} + 2 \epsilon \right) \Bigg]. \label{eqn:kl}
\end{align}
\Cref{eqn:softmax} is the usual softmax classifier loss,
but acting on our stochastic representation $z = f(x) + \epsilon$,
which is simply the output of our encoder network $f(x)$ with additive
Gaussian noise.
The $w_y$ is the $y$th row of weights in the final
linear layer outputing the logits.
$\mu_y$ are the learned class conditional means for our marginal.
$\epsilon$ are standard normal draws from an isotropic unit variance Gaussian
with the same dimension as our encoding $f(x)$.
\Cref{eqn:kl} is a stochastic sampling of the KL divergence
between our encoder likelihood and the class conditional marginal
likelihood.
$\rho$ controls the strength of the bottleneck and can vary on the whole real line.
As $\rho\to\infty$ the bottleneck is turned off.
In practice we find that $\rho$ values near but above 0 tend to work best for modest size models,
with the tendency for the best $\rho$ to approach 0 as the model
capacity increases.
Notice that in expectation the second term in the loss is $(f(x) - \mu_y)^2$, which encourages the learned
means $\mu_y$ to converge to the average of the representations
of each element in the class.
During testing we use the mean encodings and remove the
stochasticity.

In its simplest form, training a CEB classifier amounts to
injecting Gaussian random noise in the penultimate layer and
learning estimates of the class-averaged output of that layer.
In the supplemental material
we show simple modifications to the TPU-compatible
\resnet\ implementation available on GitHub from the~\citet{cloudtpuresnet}
that produce the same core \resnetfifty\ models we use for our \imagenet\
experiments.

\subsection{Consistent Classifier}
\label{sec:consist}
An alternative classifier to the standard linear layer described in~\Cref{sec:implementing} performs the Bayesian inversion on the true class-conditional marginal:
\begin{align}
p(y|z) = \frac{p(z|y)p(y)}{\sum_{y'}p(z|y')p(y')}.
\end{align}
Substituting $q(z|y)$ and using the empirical distribution over labels, we can define our variational classifier as:
\begin{align}
q(y|z) \equiv \operatorname{softmax}(q(z|y)p(y))
\end{align}
In the case that the labels are uniformly distributed, that further simplifies to $q(y|z) \equiv \operatorname{softmax}(q(z|y))$.
We call this the \textit{consistent classifier} because it is Bayes-consistent with the variational conditional marginal.
This is in contrast to the standard feed-forward classifier, which may choose to classify a region of the latent space differently from the highest density class given by the conditional marginal.

\subsection{Adversarial Attacks and Defenses}

\paragraph{Attacks.}
The first adversarial attacks were proposed in~\citet{szegedy2013intriguing,fgm}.
Since those seminal works, an enormous variety of attacks has been proposed (\citet{kurakin2016adversarial,kurakin2016adversarialml,deepfool,carliniwagner,madry2017towards,song_physical,atn}, etc.).
In this work, we will primarily consider the Projected Gradient Descent (PGD) attack~\citep{madry2017towards}, which is a multi-step variant of the early Fast Gradient Method~\citep{fgm}.
The attack can be viewed as having four parameters: $p$, the norm of the attack (typically 2 or $\infty$), $\epsilon$, the radius the the $p$-norm ball within which the attack is permitted to make changes to an input, $n$, the number of gradient steps the adversary is permitted to take, and $\epsilon_i$, the per-step limit to modifications of the current input.
In this work, we consider L$_2$ and L$_\infty$ attacks of varying $\epsilon$ and $n$, and with $\epsilon_i = \frac 4 3 \frac \epsilon n$.

\paragraph{Defenses.}
A common defense for adversarial
examples is adversarial training.
Adversarial training was originally proposed in~\citet{szegedy2013intriguing}, but was not practical until the Fast Gradient Method was introduced.
It has been studied in detail, with varied techniques~\citep{kurakin2016adversarialml,madry2017towards,madry_bugs,xie2019feature}.
Adversarial training can clearly be viewed as a form of data augmentation~\citep{tsipras2018robustness}, where instead of using some fixed set of functions to modify the training examples, we use the model itself in combination with one or more adversarial attacks to modify the training examples.
As the model changes, the distribution of modifications changes as well.
However, unlike with non-adversarial data augmentation techniques, such as AutoAugment (\autoaug) \citep{autoaugment}, adversarial training techniques considered in the literature so far cause substantial reductions in accuracy on clean test sets.
For example, the \cifar\ model described in~\citet{madry2017towards} gets 95.5\% accuracy when trained normally, but only 87.3\% when trained on L$_\infty$ adversarial examples.
More recently, \citet{xie2019feature} adversarially trains \imagenet\ models with impressive robustness to targeted PGD L$_\infty$ attacks, but at only 62.32\% accuracy on the non-adversarial test set, compared to 78.81\% accuracy for the same model trained only on clean images.

\begin{figure*}[tb]
\centering
\includegraphics[width=0.8\linewidth]{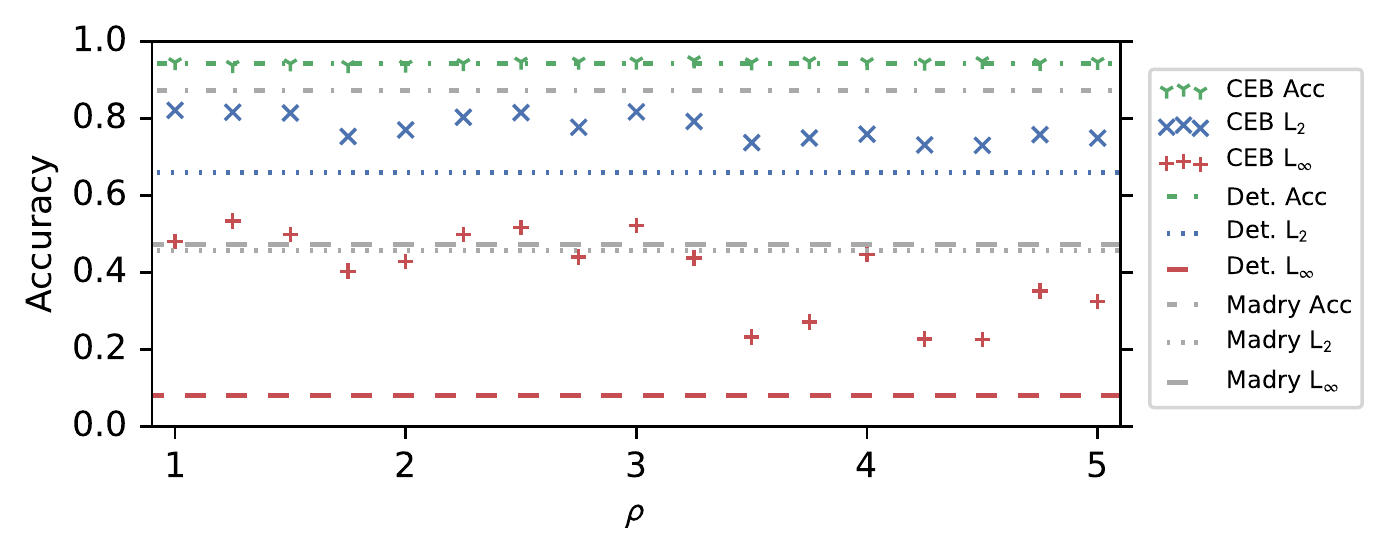}
\caption{%
  CEB $\rho$ vs. test set accuracy, and L$_2$ and L$_\infty$ PGD adversarial attacks on \cifar.
  The attack parameters were selected to be about equally difficult for the adversarially-trained WRN $28 \!\!\times\!\! 10$ model from~\citet{madry2017towards} (grey dashed and dotted lines).
  The deterministic baseline (Det.) only gets 8\% accuracy on the L$_\infty$ attacks, but gets 66\% on the L$_2$ attack, substantially better than the 45.7\% of the adversarially-trained model, which makes it clear that the adversarially-trained model failed to generalize in any reasonable way to the L$_2$ attack.
  The CEB models are always substantially more robust than Det., and many of them outperform Madry even on the L$_\infty$ attack the Madry model was trained on, but for both attacks there is a clear general trend toward more robustness as $\rho$ decreases.
  Finally, the CEB and Det. models all reach about the same accuracy, ranging from 93.9\% to 95.1\%, with Det. at 94.4\%.
  In comparison, Madry only gets 87.3\%.
  \emph{None of the CEB models is adversarially trained.}
}
\label{fig:cifar10-adv}
\end{figure*}

\begin{figure*}[tb]
\centering
\includegraphics[width=1.0\linewidth]{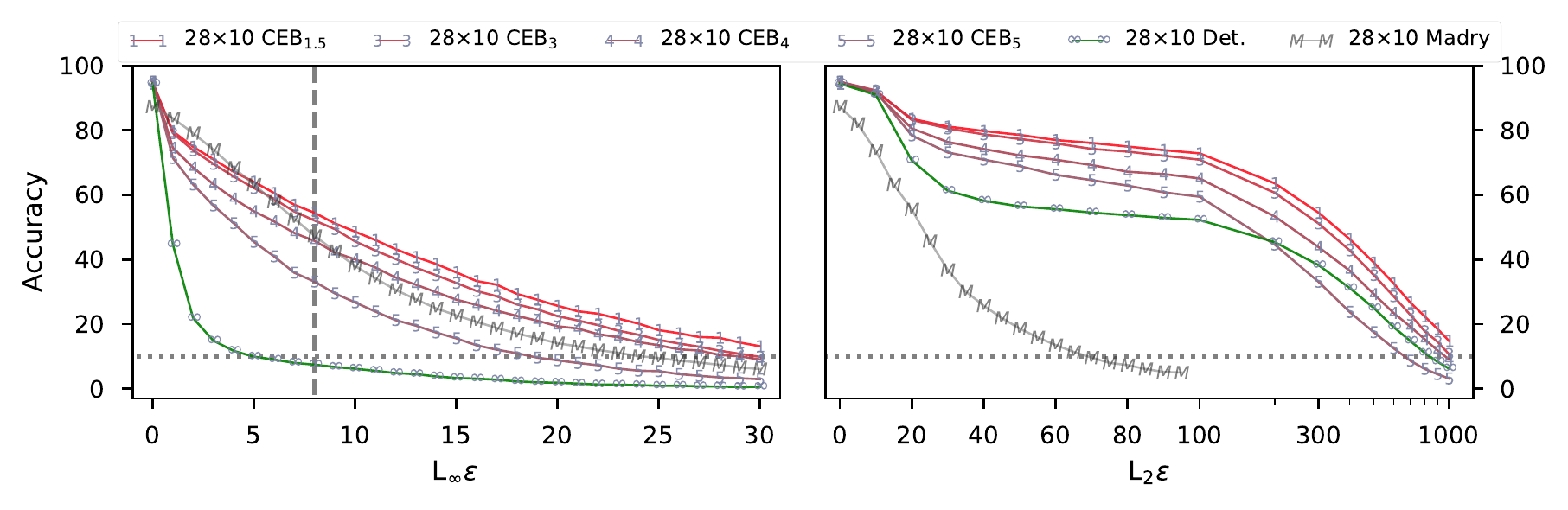}
\vspace{-0.5cm}
\caption{%
  Untargeted adversarial attacks on \cifar\ models showing both strong robustness to PGD L$_2$ and L$_\infty$ attacks, as well as good test accuracy of up to 95.1\%.
  \textbf{Left:} Accuracy on untargeted L$_\infty$ attacks at different values of $\varepsilon$ for all 10,000 test set examples.
  $28 \!\!\times\!\! 10$ indicates the Wide ResNet size.
  CEB$_x$ indicates a CEB model trained at $\rho = x$.
  Madry is the adversarially-trained model from~\citet{madry2017towards} (values provided by Aleksander Madry).
  Madry was trained with 7 steps of L$_\infty$ PGD at $\varepsilon=8$ (grey dashed line).
  All of the CEB models with $\rho \le 4$ outperform Madry across most of the values of $\epsilon$, even though they were not adversarially-trained.
  \textbf{Right:} Accuracy on untargeted L$_2$ attacks at different values of $\varepsilon$.
  Note the switch to log scale on the x axis at L$_2 \epsilon=100$.
  All values are collected at 20 steps of PGD.
  It is interesting to note that the Det. model eventually outperforms the CEB$_5$ model on L$_2$ attacks at relatively high accuracies.
  \emph{None of the CEB models is adversarially-trained.}
}
\label{fig:cifar_untargeted}
\end{figure*}

\subsection{Common Corruptions}

The Common Corruptions Benchmark~\citep{hendrycks2019benchmarking} offers a test
of model robustness to common image processing pipeline corruptions.
\Cref{fig:imagenet_summary} shows examples of the benchmark's 15 corruptions.
\imagenetc\ modifies the \imagenet\ test set with the 15 corruptions applied at five different strengths.
Within each corruption type we evaluate the average error at
each of the five levels ($E_{c} = \frac 15 \sum_{s=1}^5 E_{cs}$).
To summarize the performance across all corruptions, we report both the average corruption
error ($\text{avg} = \frac{1}{15} \sum_c E_c$) and the \emph{Mean Corruption Error} (mCE)~\citep{hendrycks2019benchmarking}:
\begin{equation}
    \text{mCE} = \frac{1}{15} \sum_c \frac{\sum_{s=1}^5 E_{cs}}{\sum_{s=1}^5 E_{cs}^\text{\alexnet}}.
\end{equation}
The mCE weights the errors on each task against the performance of a baseline \alexnet\ model.
Slightly different pipelines have been used for the \imagenetc\ task~\citep{patchgauss}.
In this work we used the \alexnet\ normalization numbers and data formulation from~\citet{fourier}.

\subsection{Natural Adversarial Examples}

The \imageneta\ Benchmark~\citep{hendrycks2019natural} is a  dataset of 7,500 naturally-occurring ``adversarial'' examples across 200 \imagenet\ classes.
The images exploit commonly-occurring weaknesses in \imagenet\ models, such as relying on textures often seen with certain class labels.

\section{Experiments}

\subsection{\cifar\ Experiments}
\label{sec:cifar}

We trained a set of 25 $28 \!\!\times\!\! 10$ Wide ResNet (WRN) CEB models on \cifar\ at $\rho \in [-1, -0.75, ..., 5]$, as well as a deterministic baseline.
They trained for 1500 epochs, lowering the learning rate by a factor of 0.3 after 500, 1000, and 1250 epochs.
This long training regime was due to our use of the original \autoaug\ policies, which requires longer training.
The only additional modification we made to the basic $28 \!\!\times\!\! 10$ WRN architecture was the removal of all Batch Normalization~\citep{batchnorm} layers.
Every small \cifar\ model we have trained with Batch Normalization enabled has had substantially worse robustness to L$_\infty$ PGD adversaries, even though typically the accuracy is much higher.
For example, $28 \!\!\times\!\! 10$ WRN CEB models rarely exceeded more than 10\% adversarial accuracy.
However, it was always still the case that lower values of $\rho$ gave higher robustness.
As a baseline comparison, a deterministic $28 \!\times\! 10$ WRN with BatchNorm, trained with \autoaug\ reaches 97.3\% accuracy on clean images, but 0\% accuracy on L$_\infty$ PGD attacks at $\epsilon = 8$ and $n=20$.
Interestingly, that model was noticeably more robust to L$_2$ PGD attacks than the deterministic baseline without BatchNorm, getting 73\% accuracy compared to 66\%.
However, it was still much weaker than the CEB models, which get over 80\% accuracy on the same attack (\Cref{fig:cifar10-adv}).
Additional training details are in the supplemental material.  

\Cref{fig:cifar10-adv} demonstrates the adversarial robustness of CEB models to
both targeted L$_2$ and L$_\infty$ attacks.
The CEB models show a marked improvement in robustness to L$_2$ attacks compared to an
adversarially-trained baseline from~\citet{madry2017towards} (denoted Madry).
\Cref{fig:cifar_untargeted} shows the robustness of five of those models to PGD attacks as $\epsilon$ is varied.
We selected the four CEB models to represent the most robust models across most of the range of $\rho$ we trained.
Note that of the 25 CEB models we trained, only the models with $\rho \ge 1$ succesfully trained.
The remainder collapsed to chance performance.
This is something we observe on all datasets when training models that are too low capacity.
Only by increasing model capacity does it become possible to train at low $\rho$.
Note that this result is predicted by the theory of the onset of learning in the Information Bottleneck and its relationship to model capacity from~\citet{iblearn}.

We additionally tested two models ($\rho=0$ and $\rho=5$) on the \cifar\ Common Corruptions test sets.
At the time of training, we were unaware that \autoaug's default policies for \cifar\ contain brightness and contrast augmentations that amount to training on those two corruptions from Common Corruptions (as mentioned in \citet{fourier}), so our results are not appropriate for direct comparison with other results in the literature.
However, they still allow us to compare the effect of bottlenecking the information between the two models.
The $\rho=5$ model reached an mCE\footnote{%
  The mCE is computed relative to a baseline model.
  We use the baseline model from~\citet{fourier}.
} of 61.2.
The $\rho=0$ model reached an mCE of 52.0, which is a dramatic relative improvement.

\begin{figure*}[p]
\centering
\includegraphics[width=0.905\linewidth]{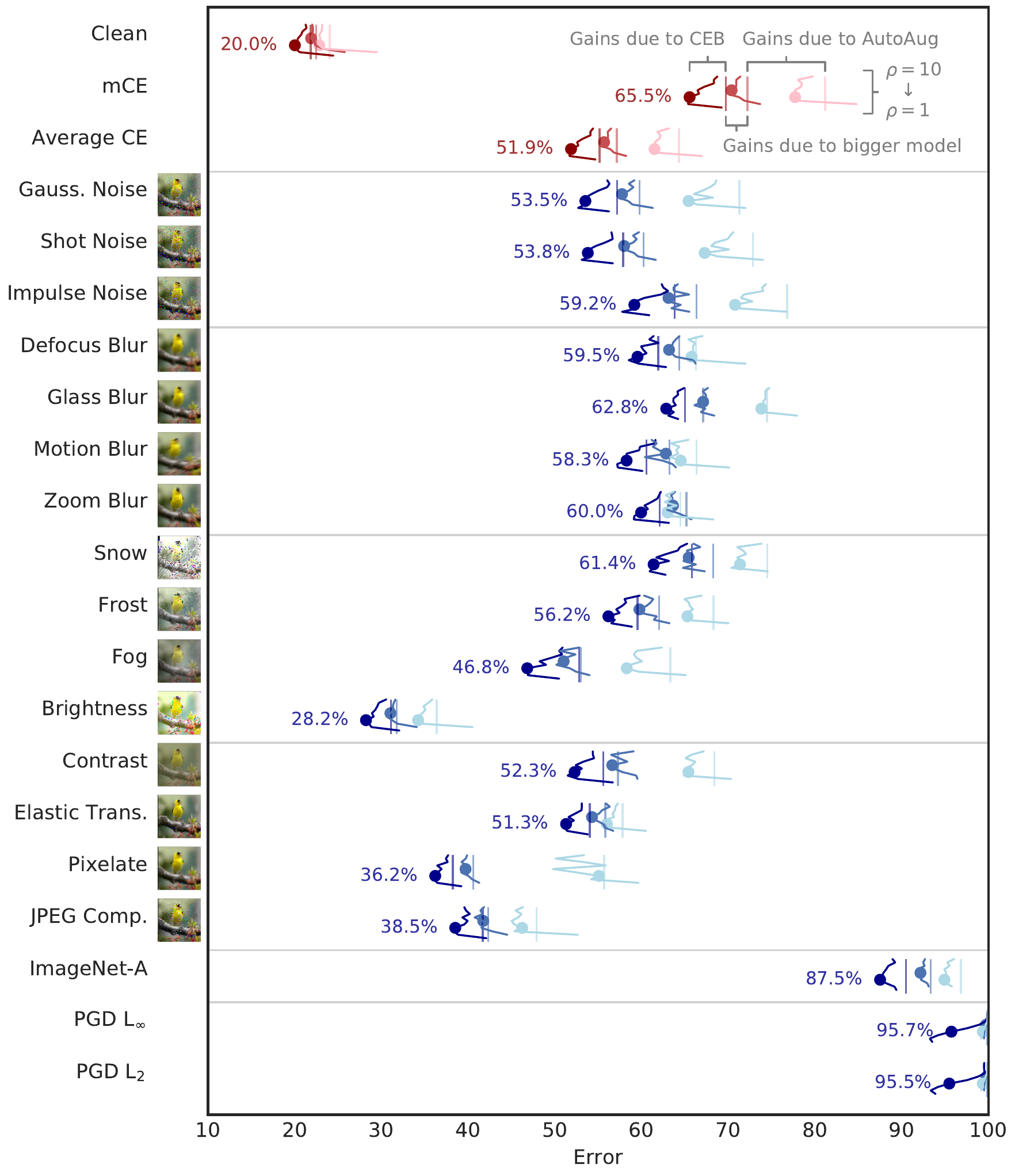}
\vspace{-1em}
\caption{
    Summary of the \resnetfifty\ \imagenetc\ experiments.
    Lower is better in all cases.
    In the main part of the figure (in blue), the average errors across corruption magnitude are shown for
    33 different networks for each of the labeled Common Corruptions, \imageneta, and targeted PGD attacks.
    The networks come in paired sets, with the vertical lines denoting the baseline XEnt network's performance,
    and then in the corresponding color the errors for each of 10 different CEB networks are shown
    with varying $\rho = [1,2, \dots, 10]$, arranged from 10 at the top to 1 at the bottom.
    The light blue lines indicate \resnetfifty\ models trained without \autoaug.
    The blue lines show the same network trained with \autoaug.
    The dark blue lines show \resnetfifty\ \autoaug\ networks that were made twice as wide.
    For these models, we display cCEB rather than CEB, which gave qualitatively similar but slightly weaker performance.
    The figure separately shows the effects of data augmentation, enlarging the model, and the additive effect of CEB on each model.
    At the top in red are shown the same data for three summary statistics.
    \texttt{clean} denotes the clean top-1 errors of each of the networks.
    \texttt{mCE} denotes the \alexnet\ regularized average corruption errors.
    \texttt{avg} shows an equally-weighted average across all common corruptions.
    The dots denote the value for each CEB network and each corruption at $\rho^*$,
    the optimum $\rho$ for the network as measured in terms of clean error.
    The values at these dots and the baseline values are given in detail in \Cref{tab:imagenet_table}.
    \Cref{fig:imagenet_summary_deep} show the same data for the \resnetdeep\ models.
    }
\label{fig:imagenet_summary}
\end{figure*}

\begin{figure*}[p]
\centering
\vspace{-3em}
\includegraphics[width=0.905\linewidth]{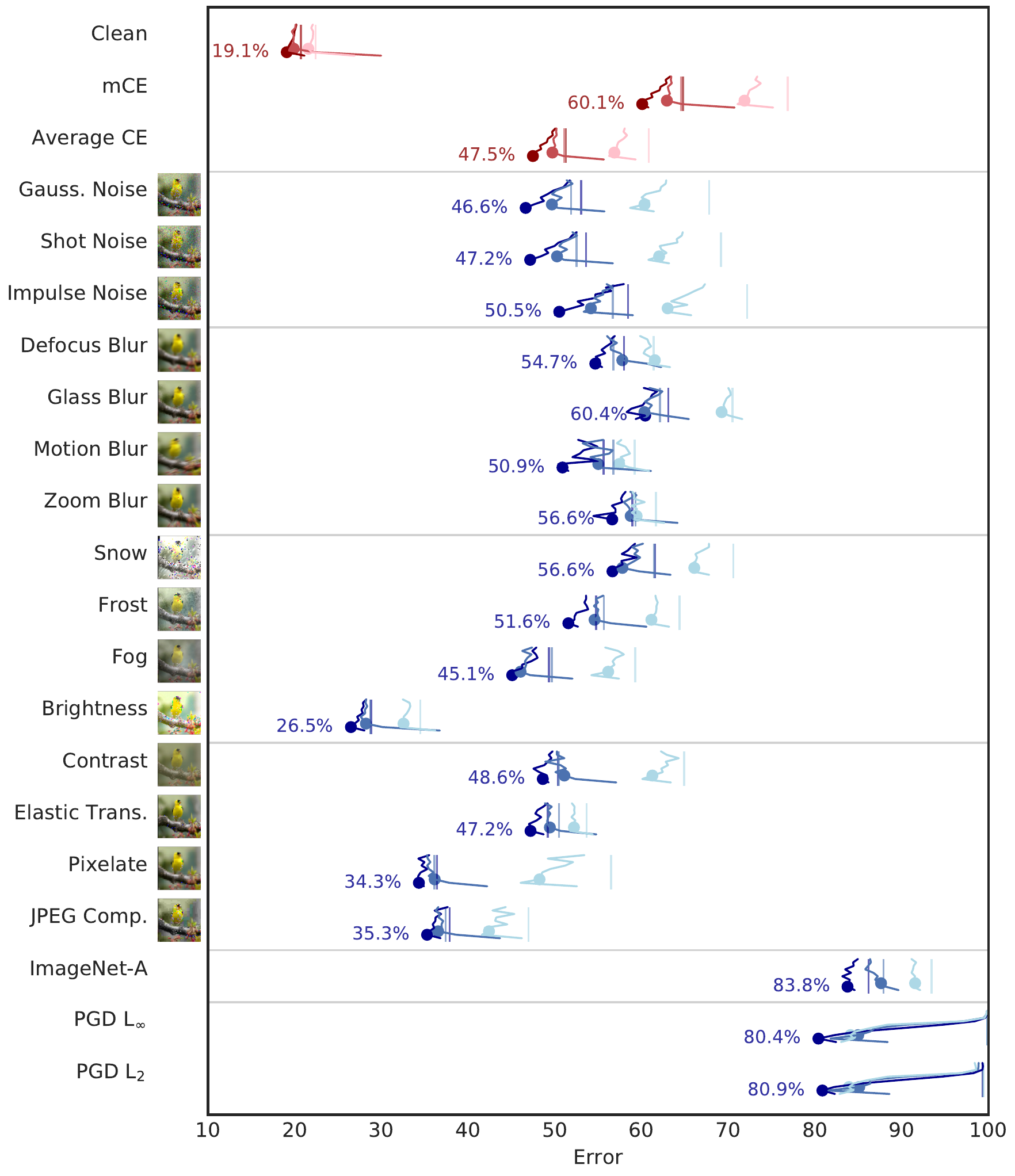}
\vspace{-1em}
\caption{
    Replication of~\Cref{fig:imagenet_summary} but for \resnetdeep.
    Lower is better in all cases.
    The light blue lines indicate \resnetdeep\ models trained without \autoaug.
    The blue lines show the same network trained with \autoaug.
    The dark blue lines show \resnetdeep\ \autoaug\ networks that were made twice as wide.
    As in~\Cref{fig:imagenet_summary}, we show the cCEB models for the largest network to reduce visual clutter.
    The deeper model shows marked improvement across the board compared to \resnetfifty,
    but the improvements due to CEB and cCEB are even more striking.
    Notice in particular the adversarial robustness to L$_\infty$ and
    L$_2$ PGD attacks for the CEB models over the XEnt baselines.
    The L$_\infty$ baselines all have error rates above $99.9\%$, so they are only barely visible along the right edge of the figure.
    See~\Cref{tab:imagenet_table} for details of the best-performing models, which correspond to the dots in this figure.
    }
\label{fig:imagenet_summary_deep}
\end{figure*}

\begin{table*}[tb]
    \setlength\tabcolsep{2.7pt} 
    \def\arraystretch{0.5}  
    \small
    \centering
    \begin{tabular}{r|ccc|cc|cc||ccc|cc|cc}
         \toprule
         Architecture & \multicolumn{3}{c|}{\resnetdeep x2} & \multicolumn{2}{c|}{\resnetdeep} & \multicolumn{2}{c||}{\resnetdeep -aa}
                      & \multicolumn{3}{c|}{\resnetfifty x2}  & \multicolumn{2}{c|}{\resnetfifty}  & \multicolumn{2}{c}{\resnetfifty -aa} \\
         Objective    & cCEB & CEB & XEnt &  CEB & XEnt &  CEB & XEnt  & cCEB & CEB & XEnt &  CEB & XEnt &  CEB & XEnt  \\
         \midrule

$\rho^*$ & 2 & 2 & NA &  3 & NA &  3 & NA &  4 & 3 & NA &  6 & NA &  4 & NA \\

Clean & \textbf{19.1\%} & 19.3\% & 20.7\% &  \textbf{19.9\%} & 20.7\% &  \textbf{21.6\%} & 22.4\% & \textbf{20.0\%} & 20.2\% & 21.8\% &  \textbf{21.9\%} & 22.5\% &  \textbf{22.8\%} & 24.0\% \\
mCE & \textbf{60.1\%} & 60.4\% & 64.8\% &  \textbf{62.9\%} & 64.6\% &  \textbf{71.9\%} & 76.9\% & \textbf{65.5\%} & 65.7\% & 69.7\% &  \textbf{70.4\%} & 72.2\% &  \textbf{77.7\%} & 81.2\% \\
Average CE & \textbf{47.5\%} & 47.8\% & 51.3\% &  \textbf{49.7\%} & 51.1\% &  \textbf{56.9\%} & 60.8\% & \textbf{51.9\%} & 52.0\% & 55.2\% &  \textbf{55.7\%} & 57.2\% &  \textbf{61.5\%} & 64.3\% \\
\midrule
Gauss. Noise & \textbf{46.6\%} & 48.0\% & 53.0\% &  \textbf{49.7\%} & 51.9\% &  \textbf{60.3\%} & 67.8\% & 53.5\% & \textbf{52.3\%} & 57.2\% &  \textbf{57.7\%} & 59.8\% &  \textbf{65.4\%} & 71.3\% \\
Shot Noise & \textbf{47.2\%} & 48.5\% & 53.6\% &  \textbf{50.3\%} & 52.5\% &  \textbf{62.0\%} & 69.2\% & 53.8\% & \textbf{52.8\%} & 57.9\% &  \textbf{58.0\%} & 60.2\% &  \textbf{67.3\%} & 72.8\% \\
Impulse Noise & \textbf{50.5\%} & 51.9\% & 58.4\% &  \textbf{54.2\%} & 56.7\% &  \textbf{63.0\%} & 72.2\% & 59.2\% & \textbf{58.3\%} & 63.8\% &  \textbf{63.1\%} & 66.4\% &  \textbf{70.8\%} & 76.8\% \\
Defocus Blur & 54.7\% & \textbf{53.8\%} & 58.0\% &  57.8\% & \textbf{56.7\%} &  61.5\% & \textbf{61.4}\% & \textbf{59.5\%} & 59.9\% & 61.9\% &  \textbf{63.2\%} & 64.4\% &  \textbf{65.8\%} & 66.3\% \\
Glass Blur & 60.4\% & \textbf{59.5\%} & 63.1\% &  \textbf{60.3\%} & 62.1\% &  \textbf{69.2\%} & 70.5\% & \textbf{62.8\%} & 64.0\% & 65.0\% &  \textbf{67.1\%} & \textbf{67.1\%} &  \textbf{73.8\%} & 74.5\% \\
Motion Blur & \textbf{50.9\%} & 52.3\% & 55.6\% &  \textbf{55.0\%} & 56.7\% &  \textbf{57.4\%} & 59.2\% & 58.3\% & \textbf{58.2\%} & 60.6\% &  \textbf{62.8\%} & 63.2\% &  \textbf{64.5\%} & 66.3\% \\
Zoom Blur & \textbf{56.6\%} & 57.0\% & 59.0\% &  \textbf{58.8\%} & 59.3\% &  \textbf{59.4\%} & 61.7\% & \textbf{60.0\%} & 60.1\% & 62.1\% &  \textbf{63.6\%} & 65.2\% &  \textbf{63.0\%} & 64.5\% \\
Snow & \textbf{56.6\%} & \textbf{56.6\%} & 61.6\% &  \textbf{57.8\%} & 61.5\% &  \textbf{66.1\%} & 70.6\% & \textbf{61.4\%} & 61.6\% & 65.8\% &  \textbf{65.4\%} & 68.3\% &  \textbf{71.4\%} & 74.5\% \\
Frost & \textbf{51.6\%} & 52.4\% & 54.8\% &  \textbf{54.6\%} & 55.7\% &  \textbf{61.1\%} & 64.4\% & \textbf{56.2\%} & \textbf{56.2\%} & 59.5\% &  \textbf{59.8\%} & 62.0\% &  \textbf{65.3\%} & 68.3\% \\
Fog & \textbf{45.1\%} & \textbf{45.1\%} & 49.3\% &  \textbf{46.0\%} & 49.6\% &  \textbf{56.2\%} & 59.3\% & \textbf{46.8\%} & 47.3\% & 52.8\% &  \textbf{51.0\%} & 53.0\% &  \textbf{58.3\%} & 63.3\% \\
Brightness & \textbf{26.5\%} & 26.6\% & 28.8\% &  \textbf{28.2\%} & 28.7\% &  \textbf{32.6\%} & 34.5\% & \textbf{28.2\%} & 28.8\% & 31.1\% &  \textbf{31.0\%} & 31.8\% &  \textbf{34.2\%} & 36.4\% \\
Contrast & 48.6\% & \textbf{48.3\%} & 50.3\% &  51.1\% & \textbf{50.5\%} &  \textbf{61.3\%} & 64.9\% & \textbf{52.3\%} & 53.5\% & 55.6\% &  \textbf{56.6\%} & 57.3\% &  \textbf{65.4\%} & 68.4\% \\
Elastic Trans. & \textbf{47.2\%} & 47.6\% & 49.2\% &  \textbf{49.4\%} & 50.5\% &  \textbf{52.2\%} & 53.7\% & \textbf{51.3\%} & 51.4\% & 54.0\% &  \textbf{54.3\%} & 55.8\% &  \textbf{56.0\%} & 57.8\% \\
Pixelate & 34.3\% & \textbf{34.2\%} & 36.4\% &  \textbf{36.1\%} & \textbf{36.1\%} &  \textbf{48.2\%} & 56.5\% & \textbf{36.2\%} & 36.3\% & 38.2\% &  \textbf{39.7\%} & 40.6\% &  \textbf{55.1\%} & 55.7\% \\
JPEG Comp. & 35.3\% & \textbf{35.1\%} & 37.9\% &  \textbf{36.5\%} & 37.4\% &  \textbf{42.4\%} & 47.0\% & \textbf{38.5\%} & 38.8\% & 41.7\% &  \textbf{41.7\%} & 42.3\% &  \textbf{46.2\%} & 47.9\% \\
\midrule
\imageneta\ & 83.8\% & \textbf{83.7\%} & 86.2\% &  \textbf{87.6\%} & 87.9\% &  \textbf{91.6\%} & 93.4\% & \textbf{87.5\%} & 88.8\% & 90.5\% &  \textbf{92.2\%} & 93.3\% &  \textbf{94.9\%} & 96.8\% \\
\midrule
PGD L$_2$ & \textbf{80.4\%} & \textbf{80.4\%} & 99.9\% &  \textbf{85.0\%} & 99.9\% &  \textbf{84.1\%} & 99.9\% & \textbf{95.7\%} & 99.4\% & 99.9\% &  \textbf{99.7\%} & 99.9\% &  \textbf{99.4\%} & 99.9\% \\
PGD L$_\infty$ & 80.9\% & \textbf{80.3\%} & 99.3\% &  \textbf{85.1\%} & 99.4\% &  \textbf{83.9\%} & 99.5\% & \textbf{95.5\%} & 99.5\% & 99.9\% &  \textbf{99.8\%} & 99.9\% &  \textbf{99.4\%} & 99.9\% \\

\bottomrule

    \end{tabular}
    \caption{
        Baseline and cross-validated CEB values for the \imagenet\ experiments.
        \textbf{cCEB} uses the consistent classifier.
        \textbf{XEnt} is the baseline cross entropy objective.
        ``\textbf{-aa}'' indicates \autoaug\ is not used during training.
        ``\textbf{x2}'' indicates the \resnet\ architecture is twice as wide.
        The CEB values reported here are denoted with the dots in \Cref{fig:imagenet_summary,fig:imagenet_summary_deep}.
        Lower values are better in all cases, and the lowest value for each architecture is shown in bold.
        For the XEnt 152x2 and 152 models, the smaller model (152) actually has better mCE and equally good top-1 accuracy, indicating that the wider model may be overfitting, but the 152x2 CEB and cCEB models substantially outperform both of them across the board.
        cCEB gives a noticeable boost over CEB for clean accuracy and mCE in both wide architectures, and large gains in the adversarial settings for all architectures except \resnet 152x2 where cCEB and CEB are essentially equally robust.
    }
    \label{tab:imagenet_table}
\end{table*}

\subsection{\imagenet\ Experiments}
\label{sec:imagenet}

To demonstrate CEB's ability to improve robustness, we trained four different \resnet\ architectures on \imagenet\ at $224 \!\times\! 224$ resolution, with and without \autoaug, using three different objective functions, and then tested them on \imagenetc, \imageneta, and targeted PGD attacks.

As a simple baseline we trained \resnetfifty\ with no data augmentation using the standard cross-entropy loss (XEnt).
We then trained the same network with CEB at ten different values of $\rho = (1, 2, \dots, 10)$.
\autoaug~\citep{autoaugment} has previously been demonstrated to improve robustness markedly on \imagenetc,
so next we trained \resnetfifty\ with \autoaug\ using XEnt.
We similarly trained these \autoaug\ \resnetfifty\ networks using CEB at the same ten values of $\rho$.
\imagenetc\ numbers are also sensitive to the model capacity.
To assess whether CEB can benefit larger models, we repeated the experiments with a modified \resnetfifty\ network where
every layer was made twice as wide, training an XEnt model and ten CEB models, all with \autoaug.
To see if there is any additional benefit or cost to using the consistent classifier (\Cref{sec:consist}), we took the same wide architecture using \autoaug\ and trained ten consistent classifier CEB (cCEB) models.
Finally, we repeated all of the previous experiments using \resnetdeep: XEnt and CEB models without \autoaug; with \autoaug; with \autoaug\ and twice as wide; and cCEB with \autoaug\ and twice as wide.
All other hyperparameters (learning rate schedule, L$_2$ weight decay scale, etc.) remained the same across all models.
All of those hyperparameters where taken from the \resnet\ hyperparameters given in the \autoaug\ paper.
In total we trained 86 \imagenet\ models: 6 deterministic XEnt models varying augmentation, width, and depth; 60 CEB models additionally varying $\rho$; and 20 cCEB models also varying $\rho$.
The results for the \resnetfifty\ models are summarized in~\Cref{fig:imagenet_summary}.
For \resnetdeep, see~\Cref{fig:imagenet_summary_deep}.
See~\Cref{tab:imagenet_table} for detailed results across the matrix of experiments.  

The CEB models highlighted in~\Cref{fig:imagenet_summary,fig:imagenet_summary_deep} and~\Cref{tab:imagenet_table} were selected by cross validation.
These were values of $\rho$ that gave the best \emph{clean} test set accuracy.
Despite being selected for classical generalization, these models also demonstrate a high degree of robustness on both average- and worst-case perturbations.
In the case that more than one model gets the same test set accuracy, we choose the model with the lower $\rho$, since we know that lower $\rho$ correlates with higher robustness.
The only model where we had to make this decision was for \resnetdeep\ with \autoaug, where five models all were within 0.1\% of each other, so we chose the $\rho=3$ model, rather than $\rho \in \{5...8\}$.

\paragraph{Accuracy, \imagenetc, and \imageneta.}
Increasing model capacity and using \autoaug\ have positive effects on classification accuracy, as well as on robustness to \imagenetc\ and \imageneta, but for all three classes of models CEB gives substantial additional improvements.
cCEB gives a small but noticeable additional gain for all three cases (except indistinguishable performance compared to CEB on \imageneta\ with the wide \resnetdeep\ architecture), indicating that enforcing variational consistency is a reasonable modification to the CEB objective.
In~\Cref{tab:imagenet_table} we can see that CEB's relative accuracy gains increase as the architecture gets larger, from gains of 1.2\% for \resnetfifty\ and \resnetdeep\ without \autoaug, to 1.6\% and 1.8\% for the wide models with \autoaug.
This indicates that even larger relative gains may be possible when using CEB to train larger architectures than those considered here.

\paragraph{Targeted PGD Attacks.}
We tested on the random-target version of the PGD L$_2$ and L$_\infty$ attacks~\citep{kurakin2016adversarial}.
Both attacks used $\epsilon=16$, $n=20$, and $\epsilon_i=2$, which is considered to be a strong attack still~\citep{xie2019feature}.
Model capacity makes a substantial difference to whitebox adversarial attacks.
In particular, none of the ResNet-50 models perform well, with all but cCEB getting less than 1\% top-1 accuracy.
However, the Resnet-152 CEB models show a dramatic improvement over the XEnt model, with top-1 accuracy increasing from 0.09\% to 17.09\% between the XEnt baseline without \autoaug\ and the corresponding $\rho=2$ CEB model, a relative increase of 187 times.
CEB and cCEB give increases nearly as large for the \autoaug\ and wide \autoaug\ models.
Interestingly, for the PGD attacks, \autoaug\ was detrimental -- the \resnetdeep\ models without \autoaug\ were more robust than those with \autoaug.
As with the accuracy results above, the relative robustness gains due to CEB increase as model capacity increases, indicating that further relative gains are possible.

\section{Conclusion}

The Conditional Entropy Bottleneck (CEB) provides a simple
mechanism to improve robustness of image classifiers.
We have shown a strong trend toward increased robustness as $\rho$ decreases in the standard $28 \!\!\times\!\! 10$ Wide \resnet\ model on \cifar, and that this increased robustness does not come at the expense of accuracy relative to the deterministic baseline.
We have shown that CEB models at a range of $\rho$ outperform an adversarially-trained baseline model, even on the attack the adversarial model was trained on, and have incidentally shown that the adversarially-trained model generalizes to at least one other attack \emph{less well} than a deterministic baseline.
Finally, we have shown that on \imagenet, CEB provides substantial gains over deterministic baselines in validation set accuracy, robustness to Common Corruptions, Natural Adversarial Examples, and targeted Projected Gradient Descent attacks.
We hope these empirical demonstrations inspire further theoretical and practical study of the use of bottlenecking techniques to encourage improvements to both classical generalization and robustness.

\clearpage
\appendix

\section{Experiment Details}

Here we give additional technical details for the \cifar\ and \imagenet\ experiments.

\subsection{\cifar\ Experiment Details}
\label{sec:cifar_details}

We trained all of the models using Adam~\citep{adam} at a base learning rate of $10^{-3}$.
We lowered the learning rate three times by a factor of 0.3 each time.
The only additional trick to train the \cifar\ models was to start with $\rho=100$, anneal down to $\rho=10$ over 2 epochs, and then anneal to the target $\rho$ over one epoch once training exceeded a threshold of 20\%.
This jump-start method is inspired by experiments on VIB in~\citet{iblearn}.
It makes it much easier to train models at low $\rho$, and appears to not negatively impact final performance.

\subsection{\imagenet\ Experiment Details}
\label{sec:imagenet_details}

We follow the learning rate schedule for the ResNet 50 from~\citet{autoaugment}, which has a top learning rate of 1.6, trains for 270 epochs, and drops the learning rate by a factor of 10 at 90, 180, and 240 epochs.
The only difference for all of our models is that we train at a batch size of 8192 rather than 4096.
Similar to the \cifar\ models, in order to ensure that the \imagenet\ models train at low $\rho$, we employ a simple jump-start.
We start at $\rho=100$ and anneal down to the target $\rho$ over 12,000 steps.
The first learning rate drop occurs a bit after 14,000 steps.
Also similar to the \cifar\ $28 \!\!\times\!\! 10$ WRN experiments, none of the models we trained at $\rho=0$ succeeded, indicating that \resnetfifty\ and wide \resnetfifty\ both have insufficient capacity to fully learn \imagenet.
We were able to train \resnetdeep\ at $\rho=0$, but only by disabling L$_2$ weight decay and using a slightly lower learning rate.
Since that involved additional hyperparameter tuning, we don't report those results here, beyond noting that it is possible, and that those models reached top-1 accuracy around 72\%.

\section{CEB Example Code}
\label{sec:ceb_code}

In~\Cref{lst:model_py,lst:main_py_head,lst:main_py} we give annotated code changes needed to make ResNet CEB models, based on the TPU-compatible ResNet implementation from the~\citet{cloudtpuresnet}.

\begin{listing*}[htbp]
\begin{minted}[fontsize=\footnotesize]{python}
# In model.py:
def resnet_v1_generator(block_fn, layers, num_classes, ...):
  def model(inputs, is_training):
    # Build the ResNet model as normal up to the following lines:
    inputs = tf.reshape(
        inputs, [-1, 2048 if block_fn is bottleneck_block else 512])
    # Now, instead of the final dense layer, just return inputs,
    # which for ResNet50 models is a [batch_size, 2048] tensor.
    return inputs
\end{minted}
\caption{Modifications to the \texttt{model.py} file.}
\label{lst:model_py}
\end{listing*}

\begin{listing*}[htbp]
\begin{minted}[fontsize=\footnotesize]{python}
# In resnet_main.py add the following imports and functions:
import tensorflow_probability as tfp
tfd = tfp.distributions

def ezx_dist(x):
  """Builds the encoder distribution, e(z|x)."""
  dist = tfd.MultivariateNormalDiag(loc=x)
  return dist

def bzy_dist(y, num_classes=1000, z_dims=2048):
  """Builds the backwards distribution, b(z|y)."""
  y_onehot = tf.one_hot(y, num_classes)
  mus = tf.layers.dense(y_onehot, z_dims, activation=None)
  dist = tfd.MultivariateNormalDiag(loc=mus)
  return dist

def cyz_dist(z, num_classes=1000):
  """Builds the classifier distribution, c(y|z)."""
  # For the classifier, we are using exactly the same dense layer
  # initialization as was used for the final layer that we removed
  # from model.py.
  logits = tf.layers.dense(
      z, num_classes, activation=None,
      kernel_initializer=tf.random_normal_initializer(stddev=.01))
  return tfd.Categorical(logits=logits)

def lerp(global_step, start_step, end_step, start_val, end_val):
  """Utility function to linearly interpolate two values."""
  interp = (tf.cast(global_step - start_step, tf.float32)
            / tf.cast(end_step - start_step, tf.float32))
  interp = tf.maximum(0.0, tf.minimum(1.0, interp))
  return start_val * (1.0 - interp) + end_val * interp
\end{minted}
\caption{Modification to the head of \texttt{resnet\_main.py}.}
\label{lst:main_py_head}
\end{listing*}

\begin{listing*}[htbp]
\begin{minted}[fontsize=\footnotesize]{python}
# Still in resnet_main.py, modify resnet_model_fn as follows:
def resnet_model_fn(features, labels, mode, params):
  # Nothing changes until after the definition of build_network:
  def build_network():
    # Elided, unchanged implementation of build_network.

  if params['precision'] == 'bfloat16':
    # build_network now returns the pre-logits, so we'll change
    # the variable name from logits to net.
    with tf.contrib.tpu.bfloat16_scope():
      net = build_network()
    net = tf.cast(net, tf.float32)
  elif params['precision'] == 'float32':
    net = build_network()

  # Get the encoder, e(z|x):
  with tf.variable_scope('ezx', reuse=tf.AUTO_REUSE):
    ezx = ezx_dist(net)
  # Get the backwards encoder, b(z|y):
  with tf.variable_scope('bzy', reuse=tf.AUTO_REUSE):
    bzy = bzy_dist(labels)

  # Only sample z during training. Otherwise, just pass through
  # the mean value of the encoder.
  if mode == tf.estimator.ModeKeys.TRAIN:
    z = ezx.sample()
  else:
    z = ezx.mean()

  # Get the classifier, c(y|z):
  with tf.variable_scope('cyz', reuse=tf.AUTO_REUSE):
    cyz = cyz_dist(z, params)

  # cyz.logits is the same as what the unmodified ResNet model would return.
  logits = cyz.logits

  # Compute the individual conditional entropies:
  hzx = -ezx.log_prob(z)  # H(Z|X)
  hzy = -bzy.log_prob(z)  # H(Z|Y) (upper bound)
  hyz = -cyz.log_prob(labels)  # H(Y|Z) (upper bound)

  # I(X;Z|Y) = -H(Z|X) + H(Z|Y)
  #          >= -hzx + hzy =: Rex, the residual information.
  rex = -hzx + hzy

  rho = 3.0  # You should make this a hyperparameter.
  rho_to_gamma = lambda rho: 1.0 / np.exp(rho)
  gamma = tf.cast(rho_to_gamma(rho), tf.float32)

  # Get the global step now, so that we can adjust rho dynamically.
  global_step = tf.train.get_global_step()

  anneal_rho = 12000  # You should make this a hyperparameter.
  if anneal_rho > 0:
    # Anneal rho from 100 down to the target rho
    # over the first anneal_rho steps.
    gamma = lerp(global_step, 0, aneal_rho,
                 rho_to_gamma(100.0), gamma)

  # Replace all the softmax cross-entropy loss computation with the following line:
  loss = tf.reduce_mean(gamma * rex + hyz)
  # The rest of resnet_model_fn can remain unchanged.
\end{minted}
\caption{Modifications to \texttt{resnet\_model\_fn} in \texttt{resnet\_main.py}.}
\label{lst:main_py}
\end{listing*}

\bibliography{bib}
\bibliographystyle{icml2020}

\end{document}